\documentclass{article}
\usepackage{spconf,amsmath,graphicx}
\usepackage{subcaption}
\usepackage{enumitem}

\usepackage{url}
\usepackage{hyperref}

\usepackage[normalem]{ulem}

\usepackage{soul}
\usepackage{xcolor}

\newcommand{\figref}[1]{Fig.~\ref{#1}}

\newcommand{\secref}[1]{Section ~\ref{#1}}

\usepackage{multirow}
\usepackage{booktabs}   
\usepackage{multirow}   
\usepackage{siunitx}    
\usepackage[table]{xcolor}
\usepackage[normalem]{ulem}

\usepackage{siunitx}

\usepackage{adjustbox}

\usepackage{threeparttable}
\usepackage{amssymb}

\title{CFE-PPAR: Compression-friendly encryption for privacy-preserving action recognition leveraging video transformers}
\twoauthors
 {$^1$ Haiwei Lin, $^2$ Shoko Imaizumi \thanks{This work was supported in part by JSPS KAKENHI under Grants JP25K07733 and JP25K07750.}}
	{$^{1, 2}$ Graduate School of Informatics \\
	Chiba University\\
	Chiba, Japan}
 {$^{3*}$ Hitoshi Kiya}
	{$^3$ Faculty of System Design\\
	Tokyo Metropolitan University\\
	Tokyo, Japan}

\begin{document}
%
\maketitle
\begin{abstract}
Privacy-preserving action recognition (PPAR) enables machines to understand human activities in videos without revealing sensitive visual content. Among the various strategies for PPAR, encryption-based methods achieve strong privacy protection while maintaining high recognition performance. However, these methods lead to a catastrophic decrease in recognition performance and visual quality when the encrypted videos are compressed. That is, the previous methods are not compression-friendly. To address these issues, in this paper, we propose the first compression-friendly encryption method for PPAR, called CFE-PPAR. In CFE-PPAR, videos encrypted with secret keys can be directly recognized by a video transformer, which uses parameters transformed by the same keys as those used for video encryption. In experiments, it is verified that CFE-PPAR outperforms previous methods on the UCF101 and HMDB51 datasets under Motion-JPEG and H.264 compression.
\end{abstract}

\section{Introduction}\label{sec:intro}
Action recognition is a fundamental task in computer vision that aims to identify and classify human activities from videos or image sequences. With the rapid development of deep neural networks (DNNs), the applications and usage environments of DNNs are becoming more widespread. In practice, action recognition systems are often deployed in cloud and edge environments. However, data transmission and storage raise serious concerns such as data privacy leakage and increasing data volume. Accordingly, privacy-preserving action recognition (PPAR) has become an urgent challenge.

PPAR can be broadly divided into two types: quality reduction-based and encryption-based. For quality reduction-based methods \cite{degradtion_0, degradtion_1, degradtion_2, degradtion_3, degradtion_4}, sensitive information in frames is obfuscated by converting it into low-fidelity representations while preserving task-relevant features; however, these methods commonly suffer from degraded recognition performance and incomplete privacy protection. In comparison, encryption-based methods \cite{enc_1, LCVE} leave scarcely any perceptible visual cues. In addition, protected content can be restored (i.e., decrypted) by authorized clients who need to conduct forensic investigations in surveillance applications. However, existing encryption-based methods do not consider applying video compression, so videos have to be transmitted in uncompressed formats.
\begin{figure}[t]
    \centering
    \includegraphics[width=1\linewidth]{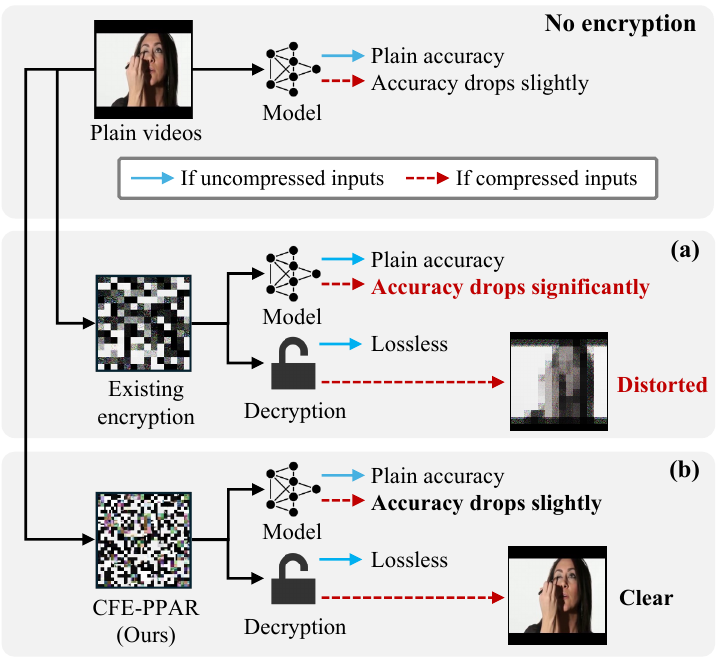}
    \caption{Utility comparison of encryption-based methods under uncompressed and compressed conditions. (a) Previous method. (b) Our CFE-PPAR.}
    \label{fig:img_abstract}
\end{figure}

Accordingly, we propose a compression-friendly encryption method for PPAR for the first time, called CFE-PPAR. As shown in \figref{fig:img_abstract} (a), when videos encrypted with existing methods are compressed \cite{enc_1, LCVE, enc_3}, recognition accuracy significantly drops, and the decryption process fails to recover faithful content. In contrast, the proposed method, CFE-PPAR, aims to address the above limitations, as shown in \figref{fig:img_abstract} (b). CFE-PPAR is carried out with a video transformer (VT) \cite{ViViT, VMAE} and two training-free components: compression-friendly encryption (CFE) and key-dependent domain adaptation (KDDA). As a result, the VT can perform accurate inference on the encrypted videos without requiring any specific fine-tuning or architectural modifications.

In experiments, the effectiveness of CFE-PPAR is verified on the UCF101 \cite{UCF101} and HMDB51 \cite{HMDB51} datasets under Motion-JPEG and H.264 compression. The experimental results show that CFE-PPAR can maintain the same accuracy as recognition on plain videos without compression and effectively mitigate catastrophic performance degradation under lossy compression.

\begin{figure*}[t]
    \centering
\includegraphics[width=1\linewidth]{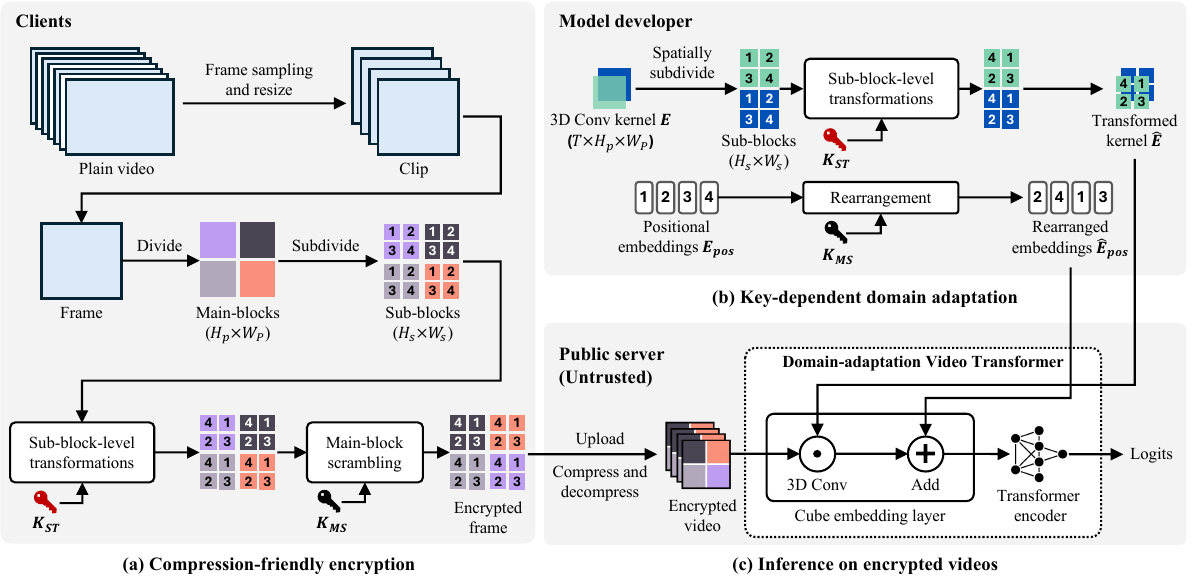}
    \caption{Overview of CFE-PPAR.}
    \label{fig:overview}
\end{figure*}

\section{Preliminary}\label{sec:pre}
\subsection{Privacy-Preserving Action Recognition}
As mentioned, methods for PPAR can be broadly categorized into quality reduction-based methods and encryption-based methods. Quality reduction-based methods, such as down-sampling \cite{degradtion_0, degradtion_1}, quantization \cite{degradtion_2}, and perturbation \cite{degradtion_3, degradtion_4}, leverage various forms of visual distortion to suppress sensitive attributes. However, these methods remain unable to sufficiently decouple sensitive attributes from task-relevant features \cite{Wu_2018_ECCV}. This inevitably compromises recognition performance and often leaves visual cues that are exploitable by reconstruction attacks.

As a promising alternative, encryption-based methods excel in recognition performance and security. Such methods have primarily explored two approaches: homomorphic encryption (HE) \cite{enc_1} and perceptual encryption (PE) \cite{LCVE}. HE \cite{enc_1} enables inference to be directly performed on encrypted data that is compatible with arithmetic computations. However, HE-based methods are computationally intensive and rely on specialized DNNs.

In contrast, \cite{LCVE} is a PE-based method, which uses key-dependent pixel shuffling to encrypt videos. While \cite{LCVE} does not match the provable security of HE, it can fully preserve recognition performance on encrypted videos without incurring additional latency or requiring architectural modifications to the model. Nevertheless, the method does not focus on compressing video data. Specifically, inference accuracy degrades significantly when encrypted data undergo compression. Moreover, once subjected to compression, the encrypted data cannot be faithfully recovered through decryption.

\subsection{Video Transformer}\label{sec:video_transformer}
Our methodology is built upon VTs \cite{ViViT, VMAE}, which have recently emerged as the dominant architecture for video understanding. For simplicity, we consider a standard VT consisting of a cube embedding layer and a vanilla transformer encoder \cite{ViT}. This model handles video inputs according to the pipeline described below.

Given a video clip $X$, the cube embedding layer divides $X$ into patches with a spatial resolution of $H_p \times W_p$ pixels. For every $T$ consecutive frames, the patches with identical spatial coordinates are stacked to form a cube with dimensions of $T \times H_p \times W_p$. Subsequently, all cubes are linearly projected into embedding vectors through 3D convolution with a kernel $E$. To retain positional information, positional embeddings $E_{pos}$ are added to the embedding vectors. Finally, the embedding vectors are passed through the transformer encoder to produce logits for action recognition.
\section{Proposed Method}\label{sec:method}
The proposed method, CFE-PPAR, is described below. 

\subsection{Overview and Threat Model}\label{sec:method_1}
\figref{fig:overview} presents an overview of the proposed method. Our scenario comprises three types of entities: clients, a model developer, and a public server, where the public server is untrusted.

As shown in \figref{fig:overview}, clients generate encrypted videos and send them to the public server after compressing the encrypted data. A block-wise encryption, which will be explained in \secref{sec:method_2} in detail, is carried out using secret keys $K_{ST}$ and $K_{MS}$. The keys are pre-shared between the authorized clients and the model developer through secure channels.

The model developer prepares a domain-adaptation VT model in which a portion of the model parameters in a trained VT are transformed with keys $K_{ST}$ and $K_{MS}$ as in \cite{key_1}. The public server decompresses the received data and inputs the encrypted data into the domain-adaptation VT to recognize the videos uploaded by the clients. Note that the public server has access to neither the secret keys nor the visual information of the videos.

A threat model includes a set of assumptions, such as an attacker's goals, knowledge, and capabilities. The aim of an adversary is to restore visual information from encrypted data. We assume that the attacker can access the encrypted data and the encryption algorithm but does not have the keys. Accordingly, the attacker can only perform ciphertext-only attacks (COAs). The proposed method is discussed by taking these environments into consideration in this paper, similar to still images.

\subsection{Compression-Friendly Encryption}\label{sec:method_2}
As described in \secref{sec:method_1}, clients encrypt video inputs, and the encrypted videos are compression-friendly, but those encrypted with existing encryption methods are not. Accordingly, we propose an encryption method for PPAR that achieves high recognition accuracy in addition to excellent compression characteristics. 

As illustrated in \figref{fig:overview} (a), the proposed method starts with preprocessing such as frame sampling and frame size conversion of original videos to align the dimensions of query videos with those of the VT on the server. After that, clients carry out the following steps to generate compression-friendly encrypted videos.
\begin{enumerate}[
    label=\textbf{A. \arabic*},
    itemsep=0pt,
    parsep=0pt,
    topsep=2pt
]
    \item Divide each resized frame into main-blocks (MBs) with a size of $H_p \times W_p$ pixels, which is equivalent to the patch size used for cubes in VTs.
    \item Subdivide each MB into sub-blocks (SBs) with a resolution $H_s \times W_s$.
    \item Randomly apply five sub-block-level transformations to each SB using key $K_{ST}$: rotation, flipping, negative-positive inversion of pixel values, RGB-channel shuffling, and sub-block scrambling (permutation).
    \item Carry out main-block scrambling (block permutation) within each frame where the scrambling order of MBs is determined by key $K_{MS}$.
\end{enumerate}

As shown in \figref{fig:samples}, frames encrypted with the above steps can protect the visual information of original videos. In addition, the blocks in encrypted videos still maintain a high correlation, so the encrypted data is compressible.  

\subsection{Key-Dependent Domain Adaptation}\label{sec:method_3}
To prevent the encryption from influencing the recognition accuracy, we apply a key-dependent domain adaptation technique to models trained with plain videos as proposed in \cite{key_1}. As shown in \figref{fig:overview} (b), the parameters of the cube embedding layer in VT are transformed by modifying the 3D convolution kernel $E$ and the positional embeddings $E_{pos}$. The procedure is given below.
\begin{enumerate}[
    label=\textbf{B. \arabic*},
    itemsep=0pt,
    parsep=0pt,
    topsep=2pt
]
    \item Subdivide the spatial components of $E$ into sub-blocks with a size of $H_s \times W_s$, which is the same size as that used in \secref{sec:method_2}.
    \item To avoid the influence of A.3, apply the five sub-block-level transformations used in A.3 to 3D convolution kernel $E$ with key $K_{ST}$.
    \item To avoid the influence of A.4, modify the positional embeddings $E_{pos}$ with key $K_{MS}$.
\end{enumerate}

The modified models, called domain-adaptation VTs, are used for action recognition on the public server.
\definecolor{acccol}{gray}{1}
\definecolor{psnrcol}{RGB}{255,255,255}

\newcommand{\accshade}[1]{\cellcolor{acccol}{#1}}
\newcommand{\badacc}[1]{\cellcolor{acccol}{\textcolor{black}{#1}}}
\newcommand{\goodacc}[1]{\cellcolor{acccol}{\textbf{{#1}}}}
\newcommand{\psnrshade}[1]{\cellcolor{psnrcol}{#1}}

\begin{table*}[t]
\centering
\scriptsize
\renewcommand{\arraystretch}{1.1}
\setlength{\tabcolsep}{0pt}

\caption{Comparison of PPAR methods in terms of recognition accuracy and visual quality under compression. \textbf{Bold} indicates the best score for each condition, excluding Plain.}
\label{tab:res}

\begin{tabular*}{\textwidth}{
@{\extracolsep{\fill}}
l
S                      
S S S S S S            
S S S S S S            
S S S S S S            
}
\toprule

\multirow{3}{*}{\textbf{Method}}
& \multicolumn{1}{c}{\multirow{2}{*}{\textbf{Uncomp}}}
& \multicolumn{6}{c}{\textbf{MJPEG}}
& \multicolumn{6}{c}{\textbf{H.264 without inter-frame prediction}}
& \multicolumn{6}{c}{\textbf{H.264 with inter-frame prediction}} \\
\cmidrule{3-8} \cmidrule{9-14} \cmidrule{15-20}

&
& \multicolumn{2}{c}{High bitrate} & \multicolumn{2}{c}{Mid bitrate} & \multicolumn{2}{c}{Low bitrate}
& \multicolumn{2}{c}{High bitrate} & \multicolumn{2}{c}{Mid bitrate} & \multicolumn{2}{c}{Low bitrate}
& \multicolumn{2}{c}{High bitrate} & \multicolumn{2}{c}{Mid bitrate} & \multicolumn{2}{c}{Low bitrate} \\

\cmidrule{2-2}
\cmidrule{3-4} \cmidrule{5-6} \cmidrule{7-8}
\cmidrule{9-10} \cmidrule{11-12} \cmidrule{13-14}
\cmidrule{15-16} \cmidrule{17-18} \cmidrule{19-20}

& Acc
& Acc & PSNR & Acc & PSNR & Acc & PSNR
& Acc & PSNR & Acc & PSNR & Acc & PSNR
& Acc & PSNR & Acc & PSNR & Acc & PSNR \\
\midrule

\multicolumn{20}{c}{\textbf{UCF101}} \\
\midrule

Plain
& \accshade{92.92}
& \accshade{91.99} & \psnrshade{35.75} & \accshade{90.88} & \psnrshade{33.25} & \accshade{81.05} & \psnrshade{29.84}
& \accshade{89.29} & \psnrshade{35.78} & \accshade{88.43} & \psnrshade{33.61} & \accshade{87.02} & \psnrshade{30.31}
& \accshade{92.36} & \psnrshade{38.01} & \accshade{92.33} & \psnrshade{36.55} & \accshade{92.18} & \psnrshade{34.44} \\

BDQ \cite{degradtion_2}
& \badacc{82.52}
& \accshade{79.32} & \psnrshade{N/A} & \accshade{77.00} & \psnrshade{N/A} & \accshade{69.07} & \psnrshade{N/A}
& \accshade{79.11} & \psnrshade{N/A} & \accshade{77.42} & \psnrshade{N/A} & \accshade{73.09} & \psnrshade{N/A}
& \accshade{80.41} & \psnrshade{N/A} & \accshade{78.35} & \psnrshade{N/A} & \accshade{75.60} & \psnrshade{N/A} \\

LCVE \cite{LCVE}
& \accshade{92.92}
& \badacc{12.18} & \psnrshade{18.09} & \badacc{6.61} & \psnrshade{17.89} & \badacc{2.34} & \psnrshade{17.39}
& \badacc{19.56} & \psnrshade{18.06} & \badacc{9.44} & \psnrshade{17.91} & \badacc{6.15} & \psnrshade{17.77}
& \badacc{20.06} & \psnrshade{18.20} & \badacc{12.79} & \psnrshade{18.01} & \badacc{4.50} & \psnrshade{17.63} \\

\textbf{CFE-PPAR (V1)}
& \goodacc{92.92}
& \goodacc{91.07} & \psnrshade{\textbf{32.24}}
& \goodacc{87.73} & \psnrshade{\textbf{29.88}}
& \goodacc{71.42} & \psnrshade{\textbf{26.95}}
& \goodacc{90.88} & \psnrshade{\textbf{32.50}}
& \goodacc{89.29} & \psnrshade{\textbf{30.64}}
& \goodacc{77.35} & \psnrshade{\textbf{27.56}}
& \goodacc{92.52} & \psnrshade{\textbf{36.77}}
& \goodacc{92.04} & \psnrshade{\textbf{35.12}}
& \goodacc{91.09} & \psnrshade{\textbf{32.77}} \\

\textbf{CFE-PPAR (V2)}
& \goodacc{92.92}
& \accshade{89.88} & \psnrshade{31.31} & \accshade{85.59} & \psnrshade{29.17} & \accshade{66.40} & \psnrshade{26.66}
& \accshade{90.06} & \psnrshade{36.42} & \accshade{86.33} & \psnrshade{29.64} & \accshade{70.45} & \psnrshade{26.42}
& \accshade{92.47} & \psnrshade{36.61} & \accshade{91.83} & \psnrshade{34.96} & \accshade{90.67} & \psnrshade{32.61} \\

\midrule
\multicolumn{20}{c}{\textbf{HMDB51}} \\
\midrule

Plain
& \accshade{70.26}
& \accshade{69.08} & \psnrshade{35.53} & \accshade{68.16} & \psnrshade{33.57} & \accshade{63.85} & \psnrshade{31.02}
& \accshade{69.48} & \psnrshade{36.94} & \accshade{69.15} & \psnrshade{35.45} & \accshade{68.63} & \psnrshade{33.36}
& \accshade{70.00} & \psnrshade{39.65} & \accshade{69.54} & \psnrshade{38.92} & \accshade{69.47} & \psnrshade{37.75}  \\

BDQ
& \badacc{68.23}
& \goodacc{67.52} & \psnrshade{N/A} & \goodacc{66.67} & \psnrshade{N/A} & \goodacc{64.71} & \psnrshade{N/A}
& \goodacc{67.32} & \psnrshade{N/A} & \goodacc{66.99} & \psnrshade{N/A} & \goodacc{65.10} & \psnrshade{N/A}
& \accshade{67.06} & \psnrshade{N/A} & \accshade{67.12} & \psnrshade{N/A} & \accshade{66.14} & \psnrshade{N/A}\\

LCVE
& \accshade{70.26}
& \badacc{7.84} & \psnrshade{19.17} & \badacc{4.57} & \psnrshade{18.95} & \badacc{2.48} & \psnrshade{18.74}
& \badacc{11.24} & \psnrshade{19.19} & \badacc{4.44} & \psnrshade{19.05}  & \badacc{2.94} & \psnrshade{18.77}
& \badacc{14.77} & \psnrshade{19.26} & \badacc{6.86} & \psnrshade{18.89} & \badacc{4.58}  & \psnrshade{18.72}  \\

\textbf{CFE-PPAR (V1)}
& \goodacc{70.26}
& \accshade{67.12} & \psnrshade{\textbf{32.18}}
& \accshade{64.90} & \psnrshade{\textbf{30.36}}
& \accshade{51.63} & \psnrshade{\textbf{30.21}}
& \accshade{66.54} & \psnrshade{\textbf{32.52}}
& \accshade{62.06} & \psnrshade{\textbf{30.21}}
& \accshade{49.08} & \psnrshade{\textbf{27.25}}
& \goodacc{68.76} & \psnrshade{\textbf{36.62}}
& \goodacc{68.43} & \psnrshade{\textbf{35.02}}
& \goodacc{66.73} & \psnrshade{\textbf{32.75}} \\

\textbf{CFE-PPAR (V2)}
& \goodacc{70.26}
& \accshade{64.11} & \psnrshade{31.31} & \accshade{60.20} & \psnrshade{29.85} & \accshade{46.01} & \psnrshade{26.88}
& \accshade{65.03} & \psnrshade{32.41} & \accshade{58.95} & \psnrshade{30.02}  & \accshade{47.19} & \psnrshade{27.00}
& \accshade{68.75} & \psnrshade{36.61} & \accshade{67.65} & \psnrshade{35.02} & \accshade{65.62} & \psnrshade{32.74} \\

\bottomrule
\end{tabular*}
\end{table*}

\section{Experiment}\label{sec:experiment}
\subsection{Experimental Setup}\label{sec:experiment_1}
Experiments were conducted on two benchmark datasets for action recognition, UCF101 and HMDB51, where UCF101 contains 3,783 test videos spanning 101 action classes, and HMDB51 is composed of 1,530 test videos from 51 action classes. 

\begin{figure}[t]
    \centering
    \includegraphics[width=1\linewidth]{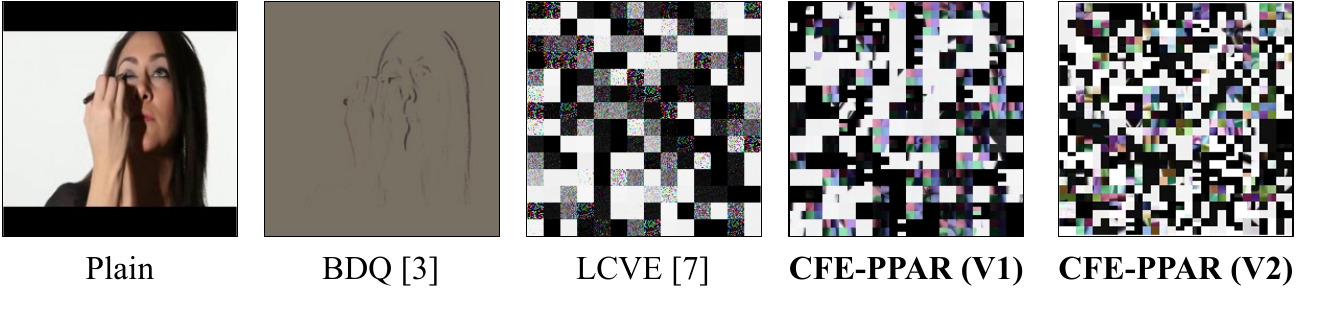}
    \caption{Frame samples of test videos produced by different PPAR methods in experiments.}
    \label{fig:samples}
\end{figure}

We used vivit-b-16x2 \cite{ViViT} as a base model in our experiments. This model is composed of a vanilla transformer encoder with a convolutional cube embedding layer using a cube size of $2 \times 16 \times16$ (i.e., $T = 2$, $H_p = 16$, and $W_p = 16$). It operates on videos with a duration of 32 frames and a spatial resolution of $224 \times 224$. To meet this input specification, we adopted uniform frame sampling and bicubic interpolation to sample and resize the frames, respectively. We initialized the model with parameters pretrained on Kinetics-400 and fine-tuned it on the target datasets. For simplicity and reproducibility, we adopted a relatively naive fine-tuning configuration without task-specific tricks such as robust training.

Our method was compared with three methods: Plain, LCVE \cite{LCVE}, and BDQ \cite{degradtion_2}, where Plain serves as a standard reference that directly takes the original videos as input and involves no modifications to the model. LCVE and BDQ are two representative current state-of-the-art methods for PPAR, where LCVE is a perceptual encryption-based method and BDQ follows a quality reduction-based strategy. BDQ trains a privacy-preserving encoder, namely BDQ encoder, to obfuscate video inputs while preserving motion features for downstream recognition. In CFE-PPAR, we set $16 \times 16$ as the MB size and $8 \times 8$ (i.e., $H_s = W_s = 8$) as the SB size. We considered two variants of CFE-PPAR, namely CFE-PPAR (V1) and CFE-PPAR (V2), according to their key assignment schemes for the MBs in a frame. Specifically, CFE-PPAR (V1) transforms all MBs using the same keys, while CFE-PPAR (V2) transforms each MB using different keys. \figref{fig:samples} shows frame samples for the methods mentioned above. 

As the experimental protocol, we first generated test videos processed by BDQ, LCVE, and CFE-PPAR, and then compressed the processed test videos and plain videos. Here, two widely used video coding methods, Motion-JPEG (MJPEG) and H.264, were used for video compression. For H.264, two prediction modes, H.264 without inter-frame prediction and H.264 with inter-frame prediction, were used for evaluation. The latter uses both intra-frame and inter-frame predictions. 

\subsection{Recognition Performance without Compression}\label{sec:experiment_2}
The experimental results are summarized in Tab. \ref{tab:res}. The accuracy of action recognition without compression (Uncomp) is given as Acc (\%) in the table. The results show that both LCVE and our method, CFE-PPAR, had the same accuracy as that of Plain, but BDQ had lower accuracy due to the use of low-quality videos. The results verified that CFE-PPAR does not cause any performance degradation when the encrypted videos are uncompressed.

\subsection{Recognition Performance with Compression}\label{sec:experiment_3}
When lossy compression is applied to videos, the action recognition accuracy degrades in general due to compression-induced noise. To evaluate the recognition performance for each method under lossy compression, we compressed videos at several bitrates as shown in Tab. \ref{tab:res}, in which High, Mid, and Low bitrates corresponded to average bitrates of 0.80, 0.60, and 0.40 bpp (bit per pixel), respectively.

LCVE showed significant accuracy degradation across all bitrates on both datasets compared with Plain. In contrast, at High and Mid bitrates, CFE-PPAR showed substantially smaller degradation than LCVE across all video coding methods. Notably, under H.264 with inter-frame prediction, CFE-PPAR remained nearly on par with Plain across all bitrates.
Although noticeable accuracy degradation was observed at Low bitrates, CFE-PPAR achieved a substantial improvement over LCVE. Both variants of CFE-PPAR exhibited similar trends in accuracy, where V1 consistently outperformed V2 by a slight margin under all conditions. While BDQ showed competitive performance on HMDB51, CFE-PPAR achieved consistently better results on UCF101. This trend indicates that CFE-PPAR scales more effectively with increasing numbers of action classes. 

\subsection{Reconstruction of Original Frames}\label{sec:experiment_4}
\begin{figure}[t]
    \centering
    \includegraphics[width=1\linewidth]{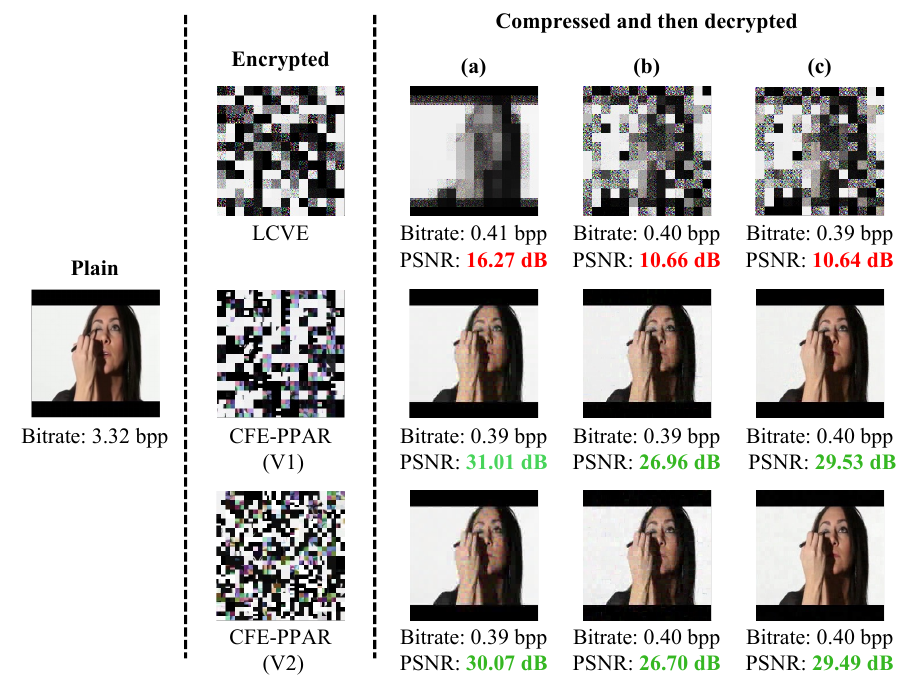}
    \caption{Decrypted frame samples from encrypted videos compressed at Low bitrate ($\approx$ 0.40 bpp). (a) MJPEG. (b) H.264 without inter-frame prediction. (c) H.264 with inter-frame prediction.}
    \label{fig:dec}
\end{figure}

PPAR methods are often required to reconstruct the visual information of original data for authorized clients. To assess the reconfigurability, we decrypted videos encrypted under different compression conditions and then calculated PSNR (peak signal-to-noise ratio) values compared with the uncompressed videos. Tab. \ref{tab:res} shows the average PSNR values of test videos for each condition, where PSNR values for BDQ were excluded since videos protected by BDQ are irreversible. From Tab. \ref{tab:res}, the videos produced by LCVE had PSNR values below 20 dB. In contrast, the videos produced by CFE-PPAR effectively avoided such catastrophic degradation. \figref{fig:dec} illustrates the effects of different video coding methods on the quality of decrypted frames. Accordingly, our method outperformed the previous PPAR methods in terms of reconfigurability. 

\begin{figure}[t]
    \centering
\includegraphics[width=1\linewidth]{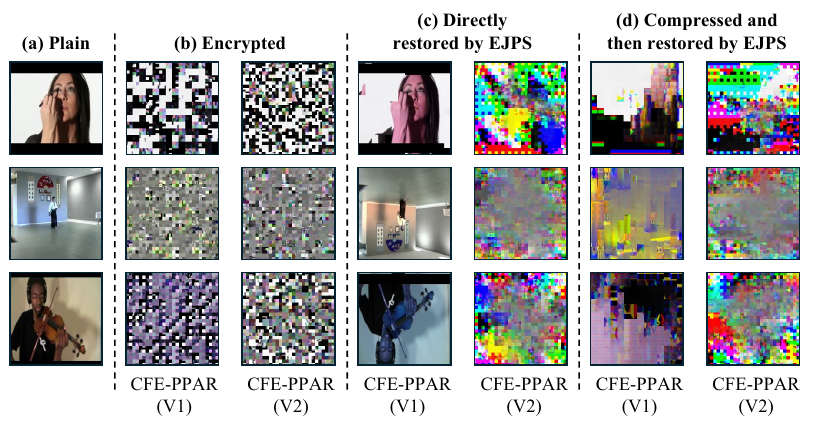}
    \caption{Encrypted videos of CFE-PPAR restored by EJPS under different compression conditions.}
    \label{fig:attk}
\end{figure}
\subsection{Attack Resistance}
As an encryption-based method, CFE-PPAR has to be robust against various attacks, including known-plaintext attacks (KPAs), chosen-plaintext attacks (CPAs), and ciphertext-only attacks (COAs). To mitigate the risks of KPAs and CPAs, CFE-PPAR adopts a one-time key policy, where each video is encrypted using a set of unique keys. In this paper, we mainly consider COAs in the threat model, which allows attackers to access the encrypted videos and the encryption algorithm, but not the secret keys.

Many studies have been conducted on the COA resistance of perceptual image encryption \cite{security_1, security_2}. However, since CFE-PPAR is a block-wise encryption method that maintains a high correlation within a block, it is potentially vulnerable to attacks based on the jigsaw puzzle solver (JPS) \cite{JPS, EJPS}. Accordingly, we adopted the extended JPS (EJPS) \cite{EJPS}, a state-of-the-art attack method that is particularly effective against encryption schemes using small block sizes.

We considered two video conditions to conduct the attack: one where the encrypted videos are not compressed, and another where the encrypted videos undergo lossy compression. Under the first condition, the videos encrypted with CFE-PPAR (V1) were almost restored by EJPS, as illustrated in \figref{fig:attk} (c). In contrast, no meaningful visual information was restored for CFE-PPAR (V2). Accordingly, CFE-PPAR (V2) was demonstrated to be more robust than CFE-PPAR (V1). For the second condition, we compressed the encrypted videos to Low bitrate ($\approx$ 0.40 bpp) using MJPEG and then applied EJPS to them. In this case, reconstructing visual information became much more difficult as illustrated in \figref{fig:attk} (d), regardless of whether the videos were encrypted with CFE-PPAR (V1) or CFE-PPAR (V2).

\section{Conclusion}
In this paper, we proposed a novel method for privacy-preserving action recognition, called CFE-PPAR. CFE-PPAR protects the visual information of videos while enabling the encrypted videos to be effectively compressed using standard compression methods such as MJPEG and H.264. Even under severe lossy compression, CFE-PPAR substantially mitigates accuracy degradation and allows authorized clients to reconstruct the original videos with reliable quality. In experiments, the effectiveness of CFE-PPAR was verified on the UCF101 and HMDB51 datasets in terms of recognition accuracy, reconstruction quality of original frames, and attack resistance.

\vfill\pagebreak

\bibliographystyle{IEEEbib}
\bibliography{strings,refs}

\end{document}